\documentclass[conference]{IEEEtran}
\IEEEoverridecommandlockouts
\usepackage{cite}
\usepackage{amsmath,amssymb,amsfonts}
\usepackage{algorithmic}
\usepackage{graphicx}
\usepackage{textcomp}
\usepackage{xcolor}
\usepackage{multirow}
\usepackage[table]{xcolor}
\usepackage{booktabs}
\usepackage{pifont}   
\usepackage{graphicx} 
\usepackage{array}
\usepackage{amsmath,amssymb}
\usepackage[colorlinks,linkcolor=black,citecolor=black,urlcolor=blue]{hyperref}
\usepackage{marvosym}
\def\BibTeX{{\rm B\kern-.05em{\sc i\kern-.025em b}\kern-.08em
    T\kern-.1667em\lower.7ex\hbox{E}\kern-.125emX}}
\usepackage{bm} 
\definecolor{mygreen}{RGB}{0,140,0}
\newcommand{\best}[1]{\textcolor{red}{\textbf{#1}}}
\newcommand{\secondbest}[1]{\textcolor{blue}{\textbf{#1}}}
\newcommand{\thirdbest}[1]{\textcolor{mygreen}{\textbf{#1}}}

\def\BibTeX{{\rm B\kern-.05em{\sc i\kern-.025em b}\kern-.08em
    T\kern-.1667em\lower.7ex\hbox{E}\kern-.125emX}}
\begin{document}

\title{ORSIFlow: Saliency-Guided Rectified Flow for Optical Remote Sensing Salient Object Detection}

\author{
    \IEEEauthorblockN{
        Haojing Chen\textsuperscript{1*},
        Zhihang Liu\textsuperscript{1*},
        Yutong Li\textsuperscript{1,2*},
        Tao Tan\textsuperscript{1},
        Haoyu Bian\textsuperscript{1},
        Qiuju Ma\textsuperscript{3\Letter}
    }
    \IEEEauthorblockA{
        \textsuperscript{1}University of Electronic Science and Technology of China, Chengdu, China
    }
    \IEEEauthorblockA{
        \textsuperscript{2}Hainan University, Haikou, China
    }
    \IEEEauthorblockA{
        \textsuperscript{3}China University of Mining \& Technology (Beijing), Beijing, China
    }

    \thanks{* Equal contribution to this work.}
    \thanks{\Letter~Corresponding author: ma200609@126.com}
    \thanks{This work was supported by the Fundamental Research Funds for the Central Universities.}
}

\maketitle

\begin{abstract}
Optical Remote Sensing Image Salient Object Detection (ORSI-SOD) remains challenging due to complex backgrounds, low contrast, irregular object shapes, and large variations in object scale. Existing discriminative methods directly regress saliency maps, while recent diffusion-based generative approaches suffer from stochastic sampling and high computational cost. In this paper, we propose ORSIFlow, a saliency-guided rectified flow framework that reformulates ORSI-SOD as a deterministic latent flow generation problem. ORSIFlow performs saliency mask generation in a compact latent space constructed by a frozen variational autoencoder, enabling efficient inference with only a few steps. To enhance saliency awareness, we design a Salient Feature Discriminator for global semantic discrimination and a Salient Feature Calibrator for precise boundary refinement. Extensive experiments on multiple public benchmarks show that ORSIFlow achieves state-of-the-art performance with significantly improved efficiency.
\end{abstract}


\section{Introduction}
\label{sec:intro}
Salient Object Detection (SOD) aims to identify the most visually salient objects in a scene and has been widely applied in many fields\cite{Yu2016_Landslide,ren2026globalscanningadaptivevisual,li2026waterflowexplicitphysicspriorrectified,xu2025dualselectivefusiontransformer}. With the increasing availability of Optical Remote Sensing Images (ORSIs) in fields including geography, agriculture, and security surveillance, Optical Remote Sensing Image Salient Object Detection (ORSI-SOD) has attracted growing research attention.

ORSIs are typically captured from high-altitude sensors and often contain complex and cluttered backgrounds, where salient objects exhibit strong visual similarity to their surroundings in terms of texture, color, or structural patterns \cite{saanet,mrbinet,Zhang2021DAFNet}. These challenges are further aggravated by irregular object topologies (e.g., meandering rivers and irregular coastlines), low object-to-background contrast, extreme scale variations, and dense small structures such as vehicles.
Most existing ORSI-SOD methods adopt a discriminative learning paradigm that directly predicts saliency maps from input images. For instance, Li \textit{et al.} \cite{Li2019NestedNet} proposed LVNet, which employs a nested architecture and dual-stream pyramids to integrate multi-level features. Li \textit{et al.} \cite{Li2023GeleNet} exploits global contextual information for ORSI-SOD into the transformer models. To enhance the capability of transformer, Zhou \textit{et al.} \cite{ERPNET}  incorporates edge-aware attention mechanism and Li \textit{et al.} \cite{Li2022AccoNet} coordinate contextual information across adjacent feature levels. Furthermore, Gu \textit{et al.} \cite{prnet} introduces feature learning to efficiently aggregate fine-grained and semantic features, while Di \textit{et al.} \cite{Di2024WeightNet} employ adaptive weighting strategies for multiscale feature aggregation.

\begin{figure}[t]
  \centering
  \includegraphics[width=\linewidth]{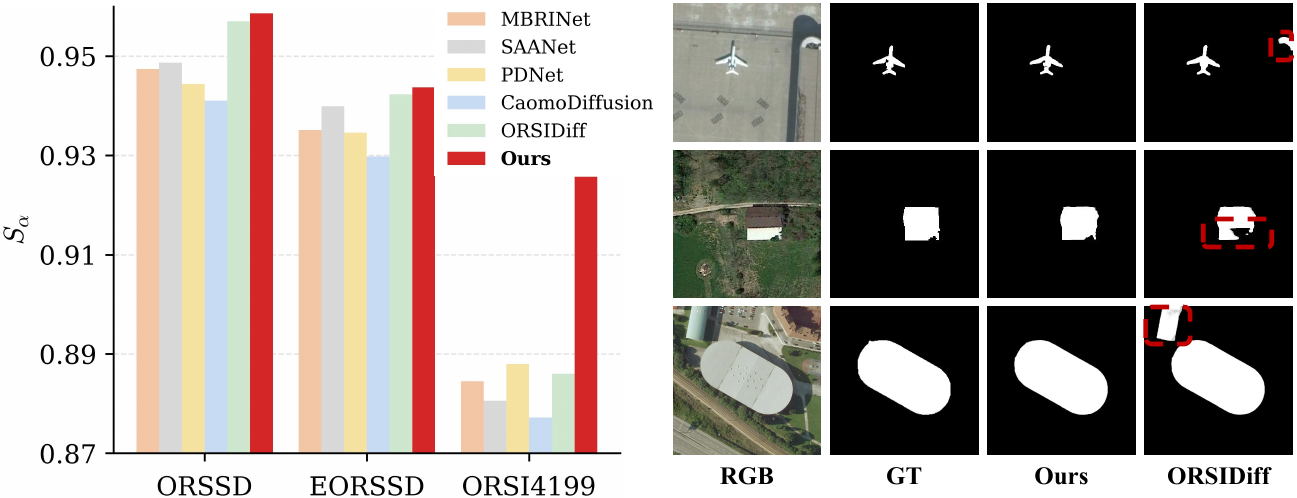}
  \caption{The left panel shows performance in terms of $S_{\alpha}$
 on three datasets, while the right panel compares our method against a previous state-of-the-art approach. Red boxes indicate segmentation errors.}
  \label{fig:head_fig}
  \vspace{-5mm}
\end{figure}

Recent studies have explored generative formulations that treat saliency detection as a mask generation problem. For example, Han \textit{et al.}\cite{orsidiff} introduces diffusion models into ORSI-SOD and achieves encouraging results by iteratively refining noisy masks. However, diffusion models typically rely on stochastic and curved denoising trajectories in the pixel space, requiring a large number of sampling steps, which limit the reasoning efficiency. As illustrated in Fig.~\ref{fig:head_fig}, such stochastic generation may still struggle to preserve fine-grained structures and small salient objects, leading to fragmented boundaries or background interference. Moreover, saliency masks exhibit relatively structured and low-entropy distributions, suggesting that a more controllable and deterministic generation process may suffice. This observation motivates us to reconsider ORSI-SOD from a flow-based perspective.

\begin{figure*}[t]
    \centering
    \includegraphics[width=\textwidth]{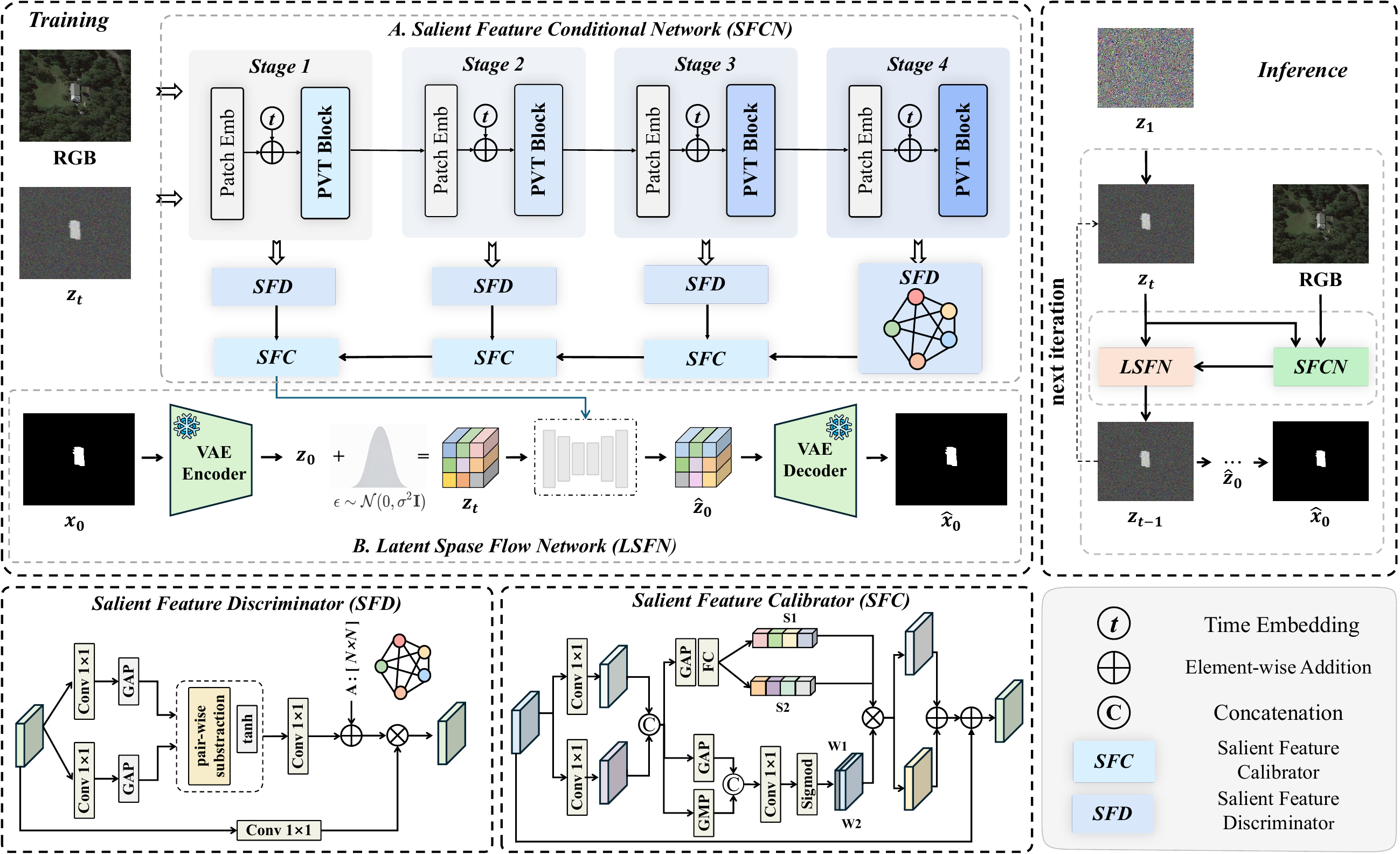}
    \caption{Overall architecture of the proposed method. The SFCN extracts multi-scale conditional features via SFD and SFC modules , while the LSFN learns a deterministic velocity field to transform the Gaussian distribution into saliency maps within the VAE latent space.}
    \label{fig:network}
    \vspace{-5mm}
\end{figure*}

In this paper, we propose ORSIFlow, which reformulates ORSI-SOD as a saliency-guided latent flow matching problem. Instead of performing generation in the pixel space, ORSIFlow conducts mask generation in a compact latent space constructed by a variational autoencoder, significantly reducing computational complexity. Built upon this latent representation, a rectified flow model learns a deterministic and approximately linear transformation between noise and saliency masks, enabling high-quality prediction with only a few inference steps. Furthermore, we introduce explicit saliency guidance through the Salient Feature Discriminator (SFD) and the Salient Feature Calibrator (SFC). Inspired by \cite{chen2021channelwisetopologyrefinementgraph} , the SFD constructs a dynamic graph topology to characterize global semantic dependencies through channel-wise reasoning. The SFC is inspired by \cite{xu2025dualselectivefusiontransformer}, which possesses the characteristic of Spatial-Spectral Selection, and is specifically designed for small objects and complex boundaries in object detection tasks. Concurrently, the SFC utilizes a dual-branch gating mechanism to dynamically aggregate local and contextual information via adaptive feature modulation. Extensive experiments on multiple public ORSI-SOD benchmarks demonstrate that ORSIFlow achieves superior performance while substantially reducing computational cost, validating the effectiveness of the proposed approach.

Our contributions are summarized as follows:
\begin{itemize}
    \item We propose ORSIFlow, a saliency-guided rectified flow framework, which performs saliency mask generation instead of direct discriminative prediction.
    
    \item We generate saliency masks in a compact latent space built by a frozen variational autoencoder and use rectified flow to map noise to masks, enabling stable training and efficient inference with few steps.
    
    \item We design two specialized guidance modules to enhance the generative quality: the Salient Feature Discriminator (SFD) effectively eliminates semantic ambiguity to maintain global consistency, while the Salient Feature Calibrator (SFC) guarantees precise boundary delineation for objects with extreme scale variations.
    
    \item Extensive experiments on multiple public ORSI-SOD benchmarks demonstrate the effectiveness of ORSIFlow.
\end{itemize}

\section{Related Work}
\textbf{ORSI-SOD.} Early ORSI-SOD methods primarily relied on CNNs, achieving progress through multi-scale feature aggregation and nested architectures\cite{Li2019NestedNet}. With the rise of self-attention, Transformers and Mamba-based modules were introduced to capture long-range dependencies and global context \cite{Li2023GeleNet}\cite{saanet}\cite{cen2025towards}. Recently, generative formulations like diffusion models have emerged to iteratively refine masks from noise \cite{RAGRNet}\cite{orsidiff}. However, these methods either lack explicit progressive refinement or suffer from the high computational cost and instability of pixel-space diffusion, often struggling to balance boundary preservation with efficient inference in complex remote sensing scenes.

\textbf{Flow‑Based Generative Learning.} Leverages continuous normalizing flows (CNFs) to model a probability flow from a simple prior to the target data distribution. Lipman et al. \cite{lipman2022flow} introduced flow matching for CNFs, learning a vector field along a fixed conditional probability path, enabling simulation‑free training and generalizing diffusion paths. Compared with diffusion models, which rely on stochastic pixel‑space denoising and incur high sampling costs, Rectified Flow \cite{liu2022flow} learns velocity fields along straight trajectories, simplifying inference and enabling few‑step sampling. 
\section{Methodology}
\subsection{Overall Architecture}
As illustrated in Fig.~2, our overall architecture consists of a Salient Feature Conditional Network and a Latent Space Flow Network. The Salient Feature Conditional Network first employs a four-stage PVTv2-b4 hierarchy to progressively extract multi-scale fused features by integrating RGB cues with the mask representation. Based on these fused features, the Salient Feature Discriminator performs relational modeling of global semantics, while the Salient Feature Calibrator corrects local details to enhance fine structures and boundary quality, yielding salient features. Then, the Latent Space Flow Network takes the salient features as conditional input and performs iterative updates in a low-dimensional latent space compressed by a VAE, finally producing the predicted saliency map.

\subsection{Salient Feature Conditional Network}
We design the Salient Feature Conditional Network (SFCN), as shown in Fig.~2(A), based on a four-stage pyramid Transformer backbone to extract hierarchical conditional features. Each stage follows a PVT-style architecture, where the spatial resolution is progressively reduced while the level of semantic abstraction is gradually increased. Given the input RGB image $I$ and the noisy latent mask $z_t$, the stage-wise embeddings are constructed as
\begin{equation}
OP_n=
\begin{cases}
\mathrm{Conv}\!\left(\mathcal{R}\!\left(\mathrm{Conv}(I)+\mathrm{Conv}(z_t)\right)\right) & n=1,\\
\mathrm{Conv}\!\left(\mathcal{R}\!\left(\mathrm{Conv}(OP_{n-1})\right)\right) & n=2,3,4.
\end{cases}
\end{equation}
where $OP_n$ denotes the stage-wise embedding at stage $n$, constructed from the RGB image $I$ and the noisy latent mask $z_t$ via convolutional projection and patch reshaping.

To adapt conditional features to different flow time steps, the continuous time variable $t \in [0,1]$ is encoded with a sinusoidal positional embedding.
\begin{equation}
T = \mathrm{SPE}(t), \quad t \sim \mathcal{U}(0,1),
\end{equation}
where $\mathrm{SPE}(\cdot)$ is the sinusoidal positional embedding.
The time embedding is injected into Transformer blocks at each stage, yielding temporally-aware multi-scale features $\{F_n\}_{n=1}^{4}$, where $F_n$ denotes the output feature of the $n$-th Transformer stage.

Given the multi-scale conditional features $\{F_n\}_{n=1}^{4}$, we apply a Salient Feature Discriminator (SFD) followed by a Salient Feature Calibrator (SFC) in a stage-wise manner to progressively refine features at different semantic levels. For the challenges in the ORSI-SOD task, where tiny objects are difficult to detect and complex boundaries are hard to determine, we propose the SFC module. Specifically, we replace the large-kernel convolution in \cite{xu2025dualselectivefusiontransformer} with a 1×1 convolution, and redesign the average pooling and max pooling operations into a dual-branch structure, enabling the model to capture finer-grained objects more effectively.

SFD aims to enhance target-background separability. We generate an importance map based on the difference between two projected channel descriptors:
\begin{equation}
    \boldsymbol{z}_1, \boldsymbol{z}_2 = \mathrm{GAP}(\mathrm{Conv}_{1\times1}(\boldsymbol{F})),
\end{equation}
\begin{equation}
    \boldsymbol{A} = \sigma(\mathrm{Conv}_{1\times1}(\tanh(\boldsymbol{z}_1 - \boldsymbol{z}_2))),
\end{equation}
where $\ominus$ denotes the pair-wise subtraction operation. Concurrently, the original feature $\boldsymbol{F}$ is mapped to a node feature representation via a third $1\times1$ convolution. We then perform graph convolution over $\mathcal{G}$ to propagate relational context across channel nodes\cite{chen2021channelwisetopologyrefinementgraph}. By incorporating an identity matrix $\boldsymbol{I}$ to retain self-loops, the graph reasoning process is formulated as:
\begin{equation}
    \boldsymbol{F}_{\mathrm{SFD}} = (\boldsymbol{A} + \boldsymbol{I}) \otimes \mathrm{Conv}_{1\times1}(\boldsymbol{F})
\end{equation}
where $\otimes$ denotes matrix multiplication.

Building upon $\boldsymbol{F}_{\mathrm{SFD}}$, SFC adopts a dual-branch projection with dual-gating fusion to improve structural fidelity in remote-sensing imagery. Specifically, $\boldsymbol{F}_{\mathrm{SFD}}$ is projected into two feature branches via separate $1\times1$ convolutions, $\boldsymbol{F}_1 = \mathrm{Conv}_{1\times1}(\boldsymbol{F}_{\mathrm{SFD}})$, with $\boldsymbol{F}_2$ obtained similarly. The two branches are then concatenated to form the fused feature:
\begin{equation}
\boldsymbol{F}_c = \mathrm{Concat}(\boldsymbol{F}_1, \boldsymbol{F}_2).
\end{equation}

We then exploit complementary global statistics from $\boldsymbol{F}_c$: average pooling captures the overall response distribution, while max pooling highlights sparse yet strong activations typical for small objects and thin boundaries,
\begin{equation}
\boldsymbol{g}_{\mathrm{avg}}=\mathrm{GAP}(\boldsymbol{F}_c), \quad\boldsymbol{g}_{\mathrm{max}}=\mathrm{GMP}(\boldsymbol{F}_c).
\end{equation}
These statistics are fused to produce dual calibration gates for the two branches,
\begin{equation}
\boldsymbol{W}_1,\boldsymbol{W}_2=\mathrm{Split}(\sigma(\mathrm{Conv}_{1\times1}(\mathrm{Concat}(\boldsymbol{g}_{\mathrm{avg}},\boldsymbol{g}_{\mathrm{max}})))).
\end{equation}
A channel-attention branch generates dual channel-wise scaling factors:
\begin{equation}
\boldsymbol{s}_1,\boldsymbol{s}_2=\mathrm{Split}(\mathrm{FC}(\mathrm{GAP}(\boldsymbol{F}_c))).
\end{equation}
Finally, SFC performs dual-gating fusion with a residual formulation:
\begin{equation}
\boldsymbol{F}_{\mathrm{cond}}=\boldsymbol{W}_1\otimes \boldsymbol{s}_1+\boldsymbol{W}_2\otimes \boldsymbol{s}_2+\boldsymbol{F}_{\mathrm{SFD}}.
\end{equation}
In remote sensing scenes, salient objects often appear at varying scales with weak boundaries. By jointly exploiting average and extreme responses with dual-branch gating, SFC enhances small-scale yet informative regions, thereby improving saliency localization accuracy.

\begin{table*}[htbp]
\centering
\caption{Quantitative comparison results in terms of $F_{\beta}^{max} $, $E_{\xi}^{max}$, $S_{\alpha}$ and $MAE$ score are presented on three benchmark datasets. In the tables, \( \uparrow \) and \( \downarrow \) indicate that higher and lower values are preferred, respectively. The top three results in each column are highlighted in red, blue, and green, respectively.
}
\resizebox{\textwidth}{!}{
\renewcommand{\arraystretch}{1}
\begin{tabular}{c|c|cccc|cccc|cccc}
\hline
\hline
\multirow{2}{*}{Method} & \multirow{2}{*}{Pub} &
\multicolumn{4}{c|}{ORSSD} & \multicolumn{4}{c|}{EORSSD} & \multicolumn{4}{c}{ORSI-4199}\\
\cline{3-14}
&&$S_{\alpha} \uparrow$ &$F_{\beta}^{max} \uparrow$ &  $E_{\xi}^{max} \uparrow$  &  $MAE\downarrow$ &
$S_{\alpha} \uparrow$ &$F_{\beta}^{max} \uparrow$ &  $E_{\xi}^{max} \uparrow$ & $MAE \downarrow$  &
$S_{\alpha} \uparrow$ &$F_{\beta}^{max} \uparrow$ &  $E_{\xi}^{max} \uparrow$ & $MAE \downarrow$
\\
\hline
GeleNet\cite{gelenet}&TIP-23&0.9328&0.9109&0.9760&0.0099&0.9324&0.8796&0.9735&0.0063&0.8730&0.8663&0.9397&0.0318\\ 

ACCoNet\cite{acconet}&TCYB-23&0.9437&0.9149&0.9796&0.0080&0.9290&0.8837&0.9727&0.0074&0.8817&0.8740&0.9478&0.0294\\

ERPNet\cite{ERPNET}&TCYB-23&0.9352&0.9036&0.9738&0.0114&0.9252&0.8743&0.9665&0.0082&0.8684&0.8610&0.9384&0.0327\\

WeightNet\cite{weightnet}&TGRS-24&
\thirdbest{0.9512}&\best{0.9278}&0.9864&0.0075&
\thirdbest{0.9417}&\thirdbest{0.8908}&\thirdbest{0.9809}&0.0060&
\thirdbest{0.8868}&\secondbest{0.8821}&0.9496&0.0290\\

PRNet\cite{prnet}&TGRS-24&
0.9425&0.9153&\thirdbest{0.9866}&0.0068&
0.9296&0.8707&\thirdbest{0.9809}&\thirdbest{0.0053}&
0.8795&0.8734&0.9491&0.0280\\

RAGRNet\cite{RAGRNet}&TGRS-24&
0.9507&0.9242&0.9861&\thirdbest{0.0066}&
0.9361&0.8852&0.9788&0.0057&
0.8859&0.8803&0.9490&0.0281\\

CamoDiff\cite{camodiff}&TPAMI-25&
0.9410&0.9116&0.9779&0.0096&
0.9298&0.8809&0.9685&0.0075&
0.8772&0.8710&0.9432&0.0312\\

MRBINet\cite{mrbinet}&TGRS-25&
0.9474&0.9199&0.9851&0.0069&
0.9351&0.8852&0.9766&0.0056&
0.8845&0.8787&0.9489&0.0287\\

SAANet\cite{saanet}&EAAI-25&
0.9487&0.9226&0.9840&0.0078&
0.9399&\secondbest{0.8918}&0.9801&0.0057&
0.8806&0.8773&0.9455&0.0302\\ 
PDNet\cite{cen2025towards}&EAAI-25&
0.9444&0.9161&0.9668&0.0080&
0.9346&0.8813&0.9524&0.0063&
\secondbest{0.8880}&0.8755&\best{0.9554}&\secondbest{0.0270}\\
ORSIDiff\cite{orsidiff}&TGRS-25&
\secondbest{0.9570}&\thirdbest{0.9255}&\secondbest{0.9880}&\secondbest{0.0054}&
\secondbest{0.9423}&0.8890&\secondbest{0.9830}&\secondbest{0.0046}&
0.8860&\thirdbest{0.8812}&\thirdbest{0.9510}&\thirdbest{0.0275}\\

\rowcolor[RGB]{255,228,227}
\textbf{Ours} & --- &
\best{0.9586}&\secondbest{0.9262}&\best{0.9883}&\best{0.0050}&
\best{0.9437}&\best{0.9010}&\best{0.9840}&\best{0.0041}&
\best{0.8935}&\best{0.8870}&\secondbest{0.9545}&\best{0.0262}\\

\hline\hline
\end{tabular}
}
\label{tab:exper}
\vspace{-3mm}
\end{table*}

\begin{figure*}[t]
  \centering
\includegraphics[width=0.95\textwidth]{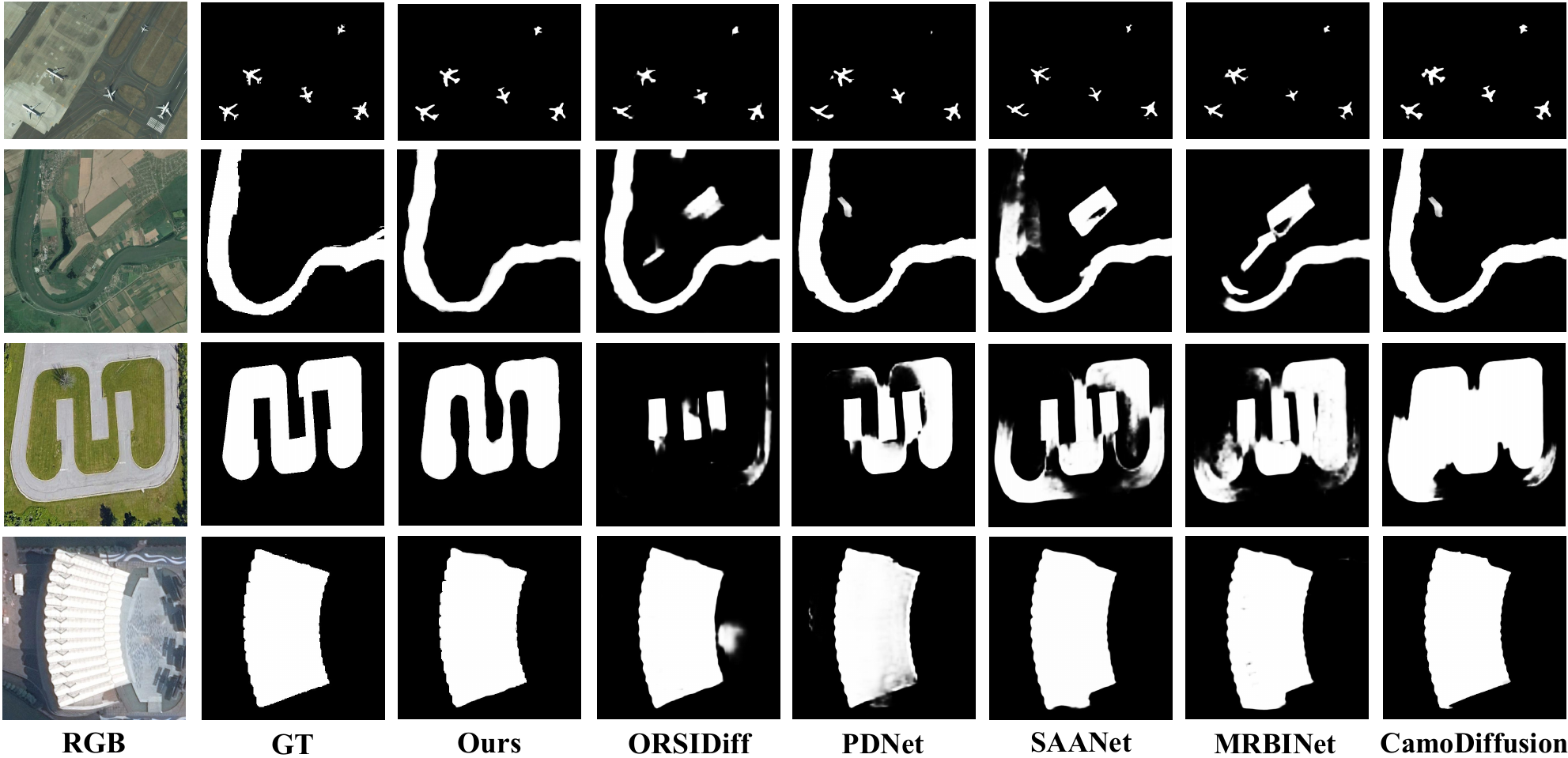}
  \caption{Qualitative comparison of ORSIFlow and other state-of-the-art methods}
  \label{fig:exper_fig}
  \vspace{-5mm}
\end{figure*}

\subsection{Latent Space Flow Network}
The Latent Space Flow Network (LSFN), as shown in Fig.~2(B), performs saliency-conditioned modeling in a compact latent space provided by a Variational Autoencoder (VAE). During the training stage, the clean saliency map $\boldsymbol{x}_0$ is mapped into the latent space through the VAE encoder:
\begin{equation}
    \boldsymbol{z}_0 = E(\boldsymbol{x}_0),
\end{equation}
where $E(\cdot)$ denotes the VAE encoder, and $\boldsymbol{z}_0$ represents the latent representation of the clean state.

Specifically, given the clean latent map $\boldsymbol{z}_0$ and a Gaussian noise sample $\boldsymbol{\epsilon} \sim \mathcal{N}(\boldsymbol{0}, \boldsymbol{I})$, the noisy state at time step $t \in [0,1]$ is defined as:
\begin{equation}
    \boldsymbol{z}_t = (1 - t)\boldsymbol{z}_0 + t\boldsymbol{\epsilon}.
\end{equation}

Conditioned on the saliency-aware embedding $\boldsymbol{F}_{\mathrm{cond}}$, LSFN adopts Rectified Flow to learn a deterministic velocity field $\boldsymbol{v}_{\theta}$ in the latent space:
\begin{equation}
    \frac{d\boldsymbol{z}_t}{dt} = \boldsymbol{v}_{\theta}(\boldsymbol{z}_t, t, \boldsymbol{F}_{\mathrm{cond}}).
\end{equation}

The network parameters are optimized by minimizing the mean squared error between the predicted velocity and the target direction $(\boldsymbol{\epsilon} - \boldsymbol{z}_0)$:
\begin{equation}
    \mathcal{L}_{\mathrm{RF}} = \mathbb{E}_{\boldsymbol{z}_0, \boldsymbol{\epsilon}, t} \left\| \boldsymbol{v}_{\theta}(\boldsymbol{z}_t, t, \boldsymbol{F}_{\mathrm{cond}}) - (\boldsymbol{\epsilon} - \boldsymbol{z}_0) \right\|_2^2.
\end{equation}

\newcommand{\cmark}{\ding{51}} 
\begin{table*}[h]
\centering
\caption{Comparison of different module ablation settings on ORSSD, EORSSD and ORSI-4199 datasets. The best result in each column is highlighted in bold.}
\label{tab:module_ablation}
\resizebox{\textwidth}{!}{
\begin{tabular}{c|ccc|cccc|cccc|cccc}
\toprule
\multirow{2}{*}{NO.} &
\multicolumn{3}{c|}{Module Ablation} &
\multicolumn{4}{c|}{ORSSD (200 test images)} &
\multicolumn{4}{c|}{EORSSD (600 test images)} &
\multicolumn{4}{c}{ORSI-4199 (2199 test images)} \\
\cmidrule(lr){2-4}
\cmidrule(lr){5-8}
\cmidrule(lr){9-12}
\cmidrule(lr){13-16}
& VAE & SFD & SFC
& $S_{\alpha}\!\uparrow$ & $F_{\beta}^{max}\!\uparrow$ & $E_{\xi}^{max}\!\uparrow$ & $MAE\!\downarrow$
& $S_{\alpha}\!\uparrow$ & $F_{\beta}^{max}\!\uparrow$ & $E_{\xi}^{max}\!\uparrow$ & $MAE\!\downarrow$
& $S_{\alpha}\!\uparrow$ & $F_{\beta}^{max}\!\uparrow$ & $E_{\xi}^{max}\!\uparrow$ & $MAE\!\downarrow$ \\
\midrule
(a) &  &  &  &
0.9453 & 0.9003 & 0.9841 & 0.0063 &
0.9303 & 0.8846 & 0.9787 & 0.0051 &
0.8696 & 0.8593 & 0.9437 & 0.0302 \\

(b) & \cmark &  &  &
0.9479 & 0.9067 & 0.9847 & 0.0061 &
0.9326 & 0.8882 & 0.9794 & 0.0050 &
0.8740 & 0.8651 & 0.9461 & 0.0297 \\

(c) &  & \cmark &  &
0.9548 & 0.9189 & 0.9863 & 0.0056 &
0.9398 & 0.8970 & 0.9821 & 0.0046 &
0.8866 & 0.8789 & 0.9515 & 0.0274 \\

(d) &  &  & \cmark &
0.9516 & 0.9128 & 0.9856 & 0.0059 &
0.9362 & 0.8923 & 0.9811 & 0.0048 &
0.8801 & 0.8724 & 0.9492 & 0.0287 \\

(e) & \cmark & \cmark &  &
0.9567 & 0.9217 & 0.9875 & 0.0052 &
0.9415 & 0.8989 & 0.9831 & 0.0044 &
0.8894 & 0.8826 & 0.9530 & 0.0271 \\

(f) & \cmark &  & \cmark &
0.9534 & 0.9150 & 0.9860 & 0.0057 &
0.9377 & 0.8941 & 0.9817 & 0.0047 &
0.8832 & 0.8752 & 0.9504 & 0.0281 \\

(g) &  & \cmark & \cmark &
0.9578 & 0.9246 & 0.9874 & 0.0052 &
0.9426 & 0.9002 & 0.9831 & 0.0043 &
0.8917 & 0.8846 & 0.9534 & 0.0268 \\

\midrule
(h) & \cmark & \cmark & \cmark &
\textbf{0.9586} & \textbf{0.9262} & \textbf{0.9883} & \textbf{0.0050} &
\textbf{0.9437} & \textbf{0.9010} & \textbf{0.9840} & \textbf{0.0041} &
\textbf{0.8935} & \textbf{0.8870} & \textbf{0.9545} & \textbf{0.0262} \\
\bottomrule
\end{tabular}
}
\vspace{-3mm}
\end{table*}

\begin{figure*}[h]
  \centering
    \includegraphics[width=\textwidth]{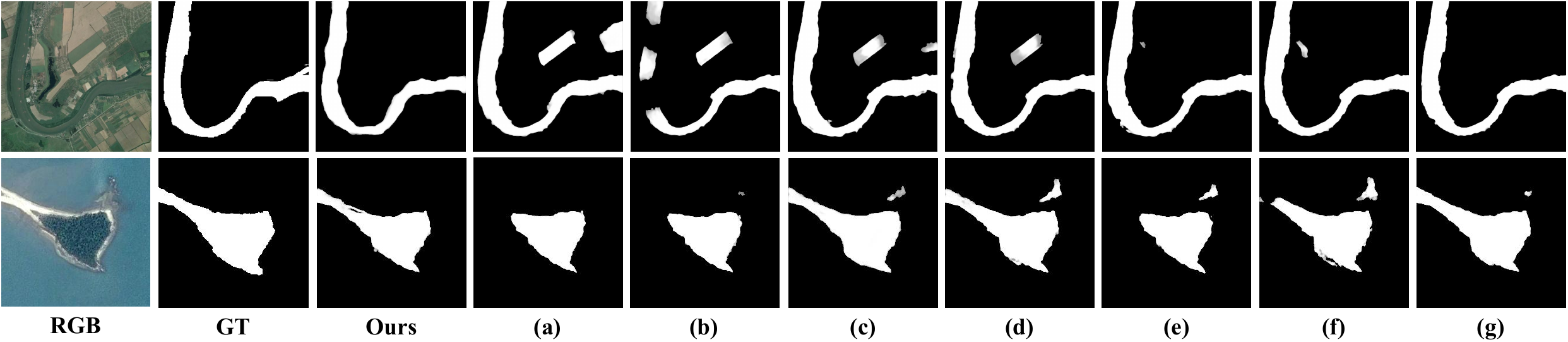}
  \caption{Visual results of the effectiveness of our modules.}
  \label{fig:abla_exper}
  \vspace{-5mm}
\end{figure*}

Finally, the refined latent representation $\hat{\boldsymbol{z}}_0$, obtained by solving the learned flow ODE backward from $t=1$ to $t=0$, is reconstructed to produce the prediction:
\begin{equation}
    \hat{\boldsymbol{x}_0} = D(\hat{\boldsymbol{z}}_0),
\end{equation}
where $D(\cdot)$ denotes the VAE decoder, and $\hat{\boldsymbol{x}_0}$ is the final predicted saliency map. Unlike the training stage, the inference stage does not require the VAE encoder. Instead, we directly initialize a Gaussian noise sample $\boldsymbol{z}_1 \sim \mathcal{N}(\boldsymbol{0}, \boldsymbol{I})$ in the latent space and feed it into the learned velocity field. By solving the learned flow ODE backward from $t=1$ to $t=0$, we obtain the clean latent representation $\hat{\boldsymbol{z}}_0$, which is then passed through the VAE decoder to generate the final clean saliency map $\hat{\boldsymbol{x}}_0$, as illustrated in Fig.~\ref{fig:network}.


\section{Experiment}

\subsection{Experimental Setup}
\textbf{Implementation Details.} The proposed model is implemented using the publicly available PyTorch framework and a single NVIDIA GeForce H100 GPU. During both the training and testing, all input images are resized to 352×352. We trained our model for 150 epochs using the AdamW optimizer with a batch size of 32 and an initial learning rate of 1e-4. 

\textbf{Datasets.} We conduct experiments on three widely-used datasets for ORSI-SOD: ORSSD~\cite{Li2019NestedNet}, EORSSD~\cite{Zhang2021DAFNet} and ORSI-4199~\cite{9511336}. The ORSSD dataset comprises 800 image pairs, with 600 pairs allocated for training and the remaining 200 for testing. The EORSSD dataset, an extended version of ORSSD, contains 2000 image pairs, of which 1400 are used for training and 600 for testing. The ORSI-4199 dataset consists of 2000 training samples and 2199 testing samples, and is widely regarded as the most challenging benchmark in ORSI-SOD due to its diverse and complex scenes.

\textbf{Evaluation Metrics.} We employed four widely used metrics to quantitatively evaluate the performance of all the methods, including S-measure ($S_a$)\cite{fan2017structure}, Mean Absolute Error (MAE), maximum E-measure ($E_\xi^{max}$)\cite{fan2018enhanced}, and maximum F-measure ($F_\beta^{max}$).

\subsection{Performance Comparison}
We conduct comparison between our proposed method and eleven other state-of-the-art (SOTA) methods, including GeleNet\cite{gelenet}, ACCoNet\cite{acconet}, ERPNet\cite{ERPNET}, WeightNet\cite{weightnet}, PRNet\cite{prnet}, RAGRNet\cite{RAGRNet}, CamoDiff\cite{camodiff}, MRBINet\cite{mrbinet}, SAANet\cite{saanet}, PDNet\cite{cen2025towards} and ORSIDiff\cite{orsidiff} To ensure a fair evaluation, the prediction results for all competing models were generated using their officially released open-source code and recommended parameter settings.

\textbf{Quantitative Comparison.} As shown in Table \ref{tab:exper}, our method consistently outperforms existing SOTA approaches on three datasets under four evaluation metrics. On ORSSD, our method achieves the highest $F_\beta^{max}$, $S_a$ and $E_\xi^{max}$, together with the lowest MAE, reflecting accurate and structurally consistent predictions. Similar trends can be observed on EORSSD, where our approach ranks first on all metrics, demonstrating its strong generalization capability. Although ORSI-4199 is more challenging due to complex backgrounds and diverse object scales, our method remains highly competitive against existing methods.

\textbf{Qualitative Comparison.} Fig.\ref{fig:exper_fig} presents qualitative comparisons between our method and several representative state-of-the-art methods. In the first row, most competing methods fail to detect small objects or miss part of the objects, whereas our method achieves more complete detections. In the second and third rows, which contain thin and elongated structures, several methods produce fragmented predictions or inaccurate boundaries, while our method better preserves the overall object structure. In the last row, where background clutter and low contrast pose additional challenges, existing approaches generate noticeable false positives, whereas our method produces cleaner segmentation results.

\subsection{Ablation Studies}

We conduct ablation studies on ORSSD, EORSSD, and ORSI-4199 to evaluate the effectiveness of each module. As reported in Table~\ref{tab:module_ablation}, introducing individual components consistently improves performance in terms of $F_\beta^{max}$, $S_a$, $E_\xi^{max}$ and MAE, while combining them leads to further gains. The full model achieves the best overall results across all datasets.
Fig. \ref{fig:exper_fig} provides qualitative comparisons, where removing certain modules results in incomplete or fragmented predictions, whereas the full model produces more accurate and coherent segmentation results.

\subsection{Efficiency Analysis}
\vspace{-3mm}
\begin{table}[h]
\centering
\caption{Efficiency comparison on the ORSI-4199 dataset with different time steps among generative ORSI-SOD methods. The best result in each column is highlighted in bold.}
\label{tab:time_steps_multi_datasets}
\renewcommand{\arraystretch}{1}
\setlength{\tabcolsep}{1.8mm}

\resizebox{\columnwidth}{!}{
    \begin{tabular}{lccc|cccc}
    \toprule
    Method & Timestep & FLOPs (G) & FPS & $S_{\alpha}\uparrow$ & $F_{\beta}^{\max}\uparrow$ & $E_{\xi}^{\max}\uparrow$ & $MAE\downarrow$ \\
    \midrule
    CamoDiff & 10 & 60.29 & 0.3 & 0.8772 & 0.8710 & 0.9432& 0.0312 \\
    ORSIDiff & 10 & 15.25 & 8   & 0.8860 & 0.8812 & 0.9510 & 0.0275\\
    Ours     & 3  & 5.68  & 19  & \textbf{0.8935} & \textbf{0.8870} & \textbf{0.9545} & \textbf{0.0262} \\
    \bottomrule
    \end{tabular}
}
\vspace{-3mm}
\end{table}

Table~\ref{tab:time_steps_multi_datasets} compares the efficiency and performance of different generative ORSI-SOD methods on the ORSI-4199 dataset. Our method achieves a superior efficiency–accuracy trade-off by using significantly fewer diffusion steps and FLOPs, while delivering the highest inference speed on GPU. Meanwhile, it attains the best overall performance in terms of $S_\alpha$, $F_\beta^{\max}$, and $E_\xi^{\max}$, along with the lowest $MAE$, demonstrating its effectiveness and computational efficiency.

\section{Conclusion}
ORSIFlow is a saliency-guided rectified flow framework for optical remote sensing salient object detection. By generating saliency masks in a compact latent space and using deterministic rectified flow, it achieves accurate predictions efficiently. The Salient Feature Discriminator and Calibrator enhance global semantic discrimination and boundary precision. Experiments on public benchmarks show ORSIFlow outperforms state-of-the-art methods with fewer inference steps, demonstrating the promise of flow-based generative modeling for robust and efficient remote sensing saliency detection.

\bibliographystyle{IEEEbib}
\bibliography{main}

@misc{ren2026globalscanningadaptivevisual,
      title={Beyond Global Scanning: Adaptive Visual State Space Modeling for Salient Object Detection in Optical Remote Sensing Images}, 
      author={Mengyu Ren and Yutong Li and Hua Li and Chuhong Wang and Runmin Cong},
      year={2026},
      eprint={2508.10542},
      archivePrefix={arXiv},
      primaryClass={cs.CV},
      url={https://arxiv.org/abs/2508.10542}, 
}

@misc{li2026waterflowexplicitphysicspriorrectified,
      title={WaterFlow: Explicit Physics-Prior Rectified Flow for Underwater Saliency Mask Generation}, 
      author={Runting Li and Shijie Lian and Hua Li and Yutong Li and Wenhui Wu and Sam Kwong},
      year={2026},
      eprint={2510.12605},
      archivePrefix={arXiv},
      primaryClass={cs.CV},
      url={https://arxiv.org/abs/2510.12605}, 
}

@article{Li2019NestedNet,
  author    = {Li, Chongyi and Cong, Runmin and Hou, Junhui and Zhang, Sanyi and Qian, Yue and Kwong, Sam},
  title     = {Nested Network With Two-Stream Pyramid for Salient Object Detection in Optical Remote Sensing Images},
  journal   = {IEEE Transactions on Geoscience and Remote Sensing},
  volume    = {57},
  number    = {11},
  pages     = {9156--9166},
  year      = {2019},
  doi       = {10.1109/TGRS.2019.2925070}
}

@article{Zhang2021DAFNet,
  author    = {Zhang, Qijian and Cong, Runmin and Li, Chongyi and Cheng, Ming{-}Ming and Fang, Yuming and Cao, Xiaochun and Zhao, Yao and Kwong, Sam},
  title     = {Dense Attention Fluid Network for Salient Object Detection in Optical Remote Sensing Images},
  journal   = {IEEE Transactions on Image Processing},
  volume    = {30},
  pages     = {1305--1317},
  year      = {2021},
  doi       = {10.1109/TIP.2020.3042084}
}

@article{Li2023GeleNet,
  author    = {Li, Gongyang and Bai, Zhen and Liu, Zhi and Zhang, Xinpeng and Ling, Haibin},
  title     = {Salient Object Detection in Optical Remote Sensing Images Driven by Transformer},
  journal   = {IEEE Transactions on Image Processing},
  volume    = {32},
  pages     = {5257--5269},
  year      = {2023},
  doi       = {10.1109/TIP.2023.3314285}
}

@article{Li2022ACCoNet,
  author    = {Li, G. and Liu, Z. and Zeng, D. and Lin, W. and Ling, H.},
  journal   = {IEEE Transactions on Cybernetics}, 
  title     = {Adjacent Context Coordination Network for Salient Object Detection in Optical Remote Sensing Images}, 
  year      = {2022},
  volume    = {53},
  number    = {1},
  pages     = {526-538},
  doi       = {10.1109/TCYB.2022.3168249}
}

@article{Di2024WeightNet,
  author    = {Di, L. and Zhang, B. and Wang, Y.},
  journal   = {IEEE Transactions on Geoscience and Remote Sensing}, 
  title     = {Multi-Scale and Multi-Dimensional Weighted Network for Salient Object Detection in Optical Remote Sensing Images}, 
  year      = {2024},
  volume    = {62},
  pages     = {1-16},
  doi       = {10.1109/TGRS.2023.3340277}
}

@ARTICLE{RAGRNet,
  author={Zhao, Jie and Jia, Yun and Ma, Lin and Yu, Lidan},
  journal={IEEE Transactions on Geoscience and Remote Sensing}, 
  title={Recurrent Adaptive Graph Reasoning Network With Region and Boundary Interaction for Salient Object Detection in Optical Remote Sensing Images}, 
  year={2024},
  volume={62},
  number={},
  pages={1-20},
  doi={10.1109/TGRS.2024.3421950}}

@inproceedings{fan2017structure,
  author={Fan, Deng-Ping and Cheng, Ming-Ming and Liu, Yun and Li, Tao and Borji, Ali},
  title={{Structure-Measure: A New Way to Evaluate Foreground Maps}},
  booktitle={Proceedings of the IEEE International Conference on Computer Vision (ICCV)},
  year={2017},
  pages={4548-4557}
}

@inproceedings{fan2018enhanced,
  author={Fan, Deng-Ping and Gong, Cheng and Cao, Yang and Ren, Bo and Cheng, Ming-Ming and Borji, Ali},
  title={{Enhanced-alignment Measure for Binary Foreground Map Evaluation}},
  booktitle={Proceedings of the Twenty-Seventh International Joint Conference on Artificial Intelligence (IJCAI)},
  year={2018},
  pages={698-704},
  doi={10.24963/ijcai.2018/97}
}

@article{Yu2016_Landslide,
  author = {Yu, Bo and Chen, Fang},
  title = {Large-scale landslide detection for practical use based on image saliency},
  journal = {Journal of Applied Remote Sensing},
  year = {2016},
  volume = {10},
  number = {4},
  pages = {045013},
  doi = {10.1117/1.JRS.10.045013},
  url = {https://doi.org/10.1117/1.JRS.10.045013}
}

@ARTICLE{9511336,

  author={Tu, Zhengzheng and Wang, Chao and Li, Chenglong and Fan, Minghao and Zhao, Haifeng and Luo, Bin},

  journal={IEEE Transactions on Geoscience and Remote Sensing}, 

  title={ORSI Salient Object Detection via Multiscale Joint Region and Boundary Model}, 

  year={2022},

  volume={60},

  number={},

  pages={1-13},

  doi={10.1109/TGRS.2021.3101359}}

@article{cen2025towards,
  title={Towards salient object detection via parallel dual-decoder network},
  author={Cen, Chaojun and Li, Fei and Li, Zhenbo and Wang, Yun},
  journal={Engineering Applications of Artificial Intelligence},
  volume={139},
  pages={109638},
  year={2025},
  publisher={Elsevier}
}

@ARTICLE{orsidiff,

  author={Han, Jinyu and Sun, Jing and Wang, Fasheng and Sun, Fuming and Li, Haojie},

  journal={IEEE Transactions on Geoscience and Remote Sensing}, 

  title={ORSIDiff: Diffusion Model for Salient Object Detection in Optical Remote Sensing Images}, 

  year={2025},

  volume={63},

  number={},

  pages={1-15},

  doi={10.1109/TGRS.2025.3579272}}

@article{saanet,
title = {Semantic awareness aggregation for salient object detection in remote sensing images},
journal = {Engineering Applications of Artificial Intelligence},
volume = {160},
pages = {111837},
year = {2025},
issn = {0952-1976},
doi = {https://doi.org/10.1016/j.engappai.2025.111837},
url = {https://www.sciencedirect.com/science/article/pii/S0952197625018391},
author = {Yanliang Ge and Taichuan Liang and Junchao Ren and Min He and Hongbo Bi and Qiao Zhang}
}

@ARTICLE{mrbinet,

  author={Jia, Yun and Zhao, Jie and Ma, Lin and Yu, Lidan},

  journal={IEEE Transactions on Geoscience and Remote Sensing}, 

  title={Multistrategy Region and Boundary Interaction Network for Salient Object Detection in Optical Remote Sensing Images}, 

  year={2025},

  volume={63},

  number={},

  pages={1-16},

  doi={10.1109/TGRS.2025.3588415}}

@ARTICLE{camodiff,

  author={Sun, Ke and Chen, Zhongxi and Lin, Xianming and Sun, Xiaoshuai and Liu, Hong and Ji, Rongrong},

  journal={IEEE Transactions on Pattern Analysis and Machine Intelligence}, 

  title={Conditional Diffusion Models for Camouflaged and Salient Object Detection}, 

  year={2025},

  volume={47},

  number={4},

  pages={2833-2848},

  doi={10.1109/TPAMI.2025.3527469}}

@ARTICLE{prnet,

  author={Gu, Shengyu and Song, Yong and Zhou, Ya and Bai, Yashuo and Yang, Xin and He, Yuxin},

  journal={IEEE Geoscience and Remote Sensing Letters}, 

  title={PRNet: Parallel Refinement Network With Group Feature Learning for Salient Object Detection in Optical Remote Sensing Images}, 

  year={2024},

  volume={21},

  number={},

  pages={1-5},

  doi={10.1109/LGRS.2024.3402821}}

@ARTICLE{weightnet,

  author={Di, Lamei and Zhang, Bin and Wang, Yiming},

  journal={IEEE Transactions on Geoscience and Remote Sensing}, 

  title={Multiscale and Multidimensional Weighted Network for Salient Object Detection in Optical Remote Sensing Images}, 

  year={2024},

  volume={62},

  number={},

  pages={1-14},

  doi={10.1109/TGRS.2024.3403268}}

@ARTICLE{acconet,

  author={Li, Gongyang and Liu, Zhi and Zeng, Dan and Lin, Weisi and Ling, Haibin},

  journal={IEEE Transactions on Cybernetics}, 

  title={Adjacent Context Coordination Network for Salient Object Detection in Optical Remote Sensing Images}, 

  year={2023},

  volume={53},

  number={1},

  pages={526-538},

  doi={10.1109/TCYB.2022.3162945}}

@article{ERPNET,
  title={Edge-Guided Recurrent Positioning Network for Salient Object Detection in Optical Remote Sensing Images},
  author={Xiaofei Zhou and Kunye Shen and Li Weng and Runmin Cong and Bolun Zheng and Jiyong Zhang and Chenggang Clarence Yan},
  journal={IEEE Transactions on Cybernetics},
  year={2022},
  volume={53},
  pages={539-552},
  url={https://api.semanticscholar.org/CorpusID:248156789}
}

@ARTICLE{gelenet,
  author={Li, Gongyang and Bai, Zhen and Liu, Zhi and Zhang, Xinpeng and Ling, Haibin},
  journal={IEEE Transactions on Image Processing}, 
  title={Salient Object Detection in Optical Remote Sensing Images Driven by Transformer}, 
  year={2023},
  volume={32},
  number={},
  pages={5257-5269},
  doi={10.1109/TIP.2023.3314285}}

@article{lipman2022flow,
  title={Flow matching for generative modeling},
  author={Lipman, Yaron and Chen, Ricky TQ and Ben-Hamu, Heli and Nickel, Maximilian and Le, Matt},
  journal={arXiv preprint arXiv:2210.02747},
  year={2022}
}

@article{liu2022flow,
  title={Flow straight and fast: Learning to generate and transfer data with rectified flow},
  author={Liu, Xingchao and Gong, Chengyue and Liu, Qiang},
  journal={arXiv preprint arXiv:2209.03003},
  year={2022}
}

@misc{xu2025dualselectivefusiontransformer,
      title={Dual Selective Fusion Transformer Network for Hyperspectral Image Classification}, 
      author={Yichu Xu and Di Wang and Lefei Zhang and Liangpei Zhang},
      year={2025},
      eprint={2410.03171},
      archivePrefix={arXiv},
      primaryClass={cs.CV},
      url={https://arxiv.org/abs/2410.03171}, 
}

@misc{chen2021channelwisetopologyrefinementgraph,
      title={Channel-wise Topology Refinement Graph Convolution for Skeleton-Based Action Recognition}, 
      author={Yuxin Chen and Ziqi Zhang and Chunfeng Yuan and Bing Li and Ying Deng and Weiming Hu},
      year={2021},
      eprint={2107.12213},
      archivePrefix={arXiv},
      primaryClass={cs.CV},
      url={https://arxiv.org/abs/2107.12213}, 
}

\end{document}